\def\year{2022}\relax
\documentclass[letterpaper]{article} %
\usepackage{aaai22}  %
\usepackage{times}  %
\usepackage{helvet}  %
\usepackage{courier}  %
\usepackage[hyphens]{url}  %
\usepackage{graphicx} %
\usepackage{natbib}  %
\usepackage{caption} %
\DeclareCaptionStyle{ruled}{labelfont=normalfont,labelsep=colon,strut=off} %
\usepackage{algorithm}
\usepackage{algorithmic}
\usepackage[switch]{lineno}

\usepackage{newfloat}
\usepackage{listings}
\floatstyle{ruled}
\newfloat{listing}{tb}{lst}{}
\floatname{listing}{Listing}

\usepackage{booktabs} %
\usepackage{fontawesome}
\usepackage{amsmath}
\usepackage{amssymb}
\usepackage{makecell}
\usepackage{tabularx}
\usepackage{xspace}
\usepackage{array}
\usepackage{comment}
\usepackage{enumitem}
\usepackage{todonotes}
\usepackage{nccmath}

\newcommand\PaperTitle{Few-Shot Cross-Lingual Stance Detection with Sentiment-Based Pre-Training}

\newcommand\XLMR{\mbox{XLM-R}\xspace}
\newcommand\XLMRb{\mbox{XLM-R}\textsubscript{Base}\xspace}

\newcommand\DatasetsCount{15\xspace}

\newcommand\fmacro{F\textsubscript{1}}

\newcolumntype{L}[1]{>{\raggedright\let\newline\\\arraybackslash\hspace{0pt}}m{#1}}
\newcolumntype{C}[1]{>{\centering\let\newline\\\arraybackslash\hspace{0pt}}m{#1}}
\newcolumntype{R}[1]{>{\raggedleft\let\newline\\\arraybackslash\hspace{0pt}}m{#1}}

\title{\PaperTitle}
\author {
    Momchil Hardalov,\textsuperscript{\rm 1,2}
    Arnav Arora,\textsuperscript{\rm 1,3}
    Preslav Nakov,\textsuperscript{\rm 1,4}
    Isabelle Augenstein\textsuperscript{\rm 1,3}
}
\affiliations {
    \textsuperscript{\rm 1} Checkstep Research\\
    \textsuperscript{\rm 2} Sofia University ``St. Kliment Ohridski'', Bulgaria\\
    \textsuperscript{\rm 3} University of Copenhagen, Denmark\\
    \textsuperscript{\rm 4} Qatar Computing Research Institute, HBKU, Doha, Qatar\\
    {\tt \{momchil, arnav, preslav.nakov, isabelle\}@checkstep.com}
}

\usepackage{bibentry}

\begin{document}

\maketitle
\begin{abstract}

The goal of stance detection is to determine the viewpoint expressed in a piece of text towards a target. These viewpoints or contexts are often expressed in many different languages depending on the user and the platform, which can be a local news outlet, a social media platform, a news forum, etc. %
Most research in stance detection, however, has been limited to working with a single language and on a few limited targets, with little work on cross-lingual stance detection. Moreover, non-English sources of labelled data are often scarce and present additional challenges.
Recently, large multilingual language models have substantially improved the performance on many non-English tasks, especially such with limited numbers of examples. This highlights the importance of model pre-training and its ability to learn from few examples. In this paper, we present the most comprehensive study of cross-lingual stance detection to date: we experiment with \DatasetsCount diverse datasets in 12 languages from 6 language families, and with 6 low-resource evaluation settings each. For our experiments, we build on pattern-exploiting training, proposing the addition of a novel label encoder to simplify the verbalisation procedure. We further propose sentiment-based generation of stance data for pre-training, which shows sizeable improvement of more than 6\% \fmacro\ absolute in low-shot settings compared to several strong baselines. %
 
\end{abstract}

\section{Introduction}
\label{sec:introduction}

As online speech gets democratised, we see an ever-growing representation of non-English languages on online platforms. %
However, in stance detection
multilingual resources are scarce~\citep{joshi-etal-2020-state}.
While English datasets exist for various domains and in different sizes, non-English and multilingual datasets are often small (under a thousand examples~\citep{lai2018conref, LAI2020101075:FRAITA,lozhnikov2020rustance}) and focus on narrow, potentially country- or culture-specific topics, such as a referendum~\citep{taule2017overview,lai2018conref}, a person~\citep{hercig2017detecting,LAI2020101075:FRAITA}, or a notable event~\citep{Swami2018AnEC}, with few exceptions~\citep{vamvas2020xstance}.

Recently, notable progress was made in zero- and few-shot learning for natural language processing (NLP) %
using pattern-based training~\citep{brown2020language,schick-schutze-2021-exploiting,gao-etal-2021-making}. These approaches shed light on the ability of pre-trained models to perform in low-resource scenarios, making them an ideal option for modelling cross-lingual stance. Yet, previous work mostly focused on single-task and single-language scenarios. In contrast, here we study their multilingual performance, and their ability to transfer knowledge across tasks and datasets. %
Moreover, a limitation of these models, especially for pattern-exploiting training~\citep{schick-schutze-2021-exploiting}, is the need for label verbalisation, i.e.,~to identify single words describing labels. This can be inconvenient for label-rich and nuanced tasks such as stance detection. We overcome this limitation by introducing a label encoder.%

Another line of research is transfer learning from different tasks and domains. Recent studies have shown that multi-task and multi-dataset learning can increase both the accuracy and the robustness of stance detection models~\citep{schiller2021stance,hardalov-etal-2021-cross}. Nonetheless, pre-training should not necessarily be performed on the same task; in fact, it is important to select the auxiliary task to pre-train on carefully~\citep{poth2021pre}. Additional or auxiliary data, albeit from a similar task, can also improve performance. An appealing candidate for stance detection is sentiment analysis, due to its semantic relationship to stance~\citep{ebrahimi-etal-2016-joint,sobhani-etal-2016-detecting}.

Our work makes the following contributions:
\begin{itemize}
    \item We present the largest study of cross-lingual stance detection, covering 15 datasets in 12 diverse languages from 6 language families.
    \item We explore the capabilities of pattern training both in a few-shot and in a full-resource cross-lingual setting.
    \item We introduce a novel label encoding mechanism to overcome the limitations of predicting multi-token labels and the need for verbalisation (single-token labels).
    \item We diverge from stance-to-stance transfer by proposing a novel semi-supervised approach to produce automatically labelled instances with a trained %
    sentiment model, leading to sizeable improvements over strong baselines.
    \item We show that our newly introduced semi-supervised approach outperforms models fine-tuned on few shots from multiple cross-lingual datasets, while being competitive with pre-trained models on English stance datasets.
\end{itemize}
\section{Method}
\label{sec:approach}

We propose an end-to-end few-shot learning, and a novel noisy sentiment-based stance detection pre-training.

\begin{figure}[!t]%
    \centering
    \includegraphics[width=\columnwidth]{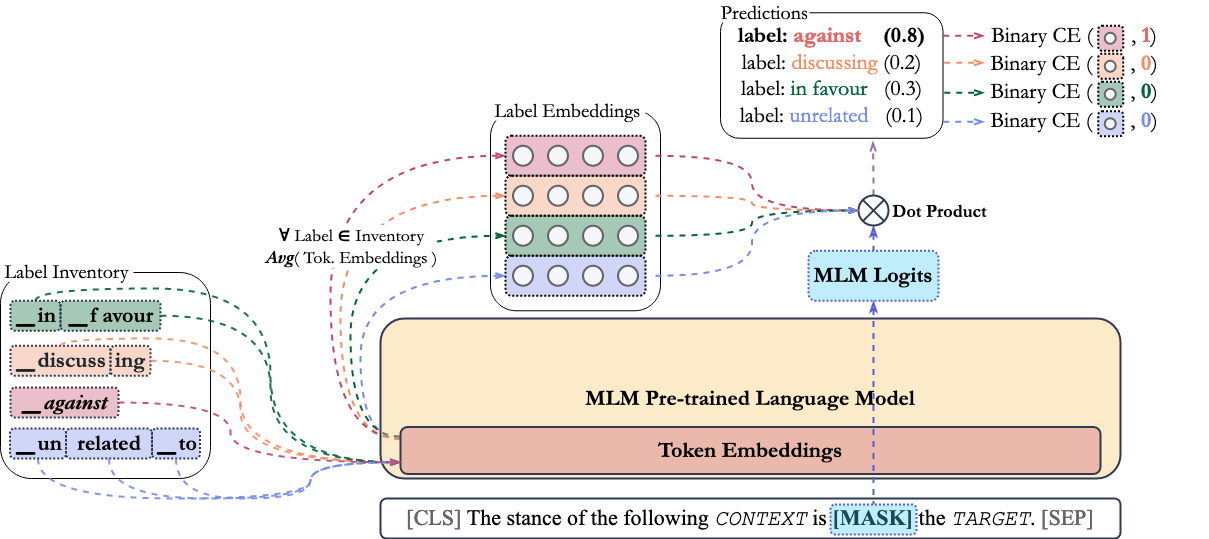}
    \caption{The architecture of the proposed method and the prompt used for prediction. The \emph{CONTEXT} and the \emph{TARGET} are replaced with the corresponding ones for each example. The label inventory comes from the training dataset.}
    \label{fig:approach}
\end{figure}

\subsection{Few-Shot Pattern-Exploiting Learning (PET)}

PET and its variants~\cite{schick-schutze-2021-exploiting,schick-schutze-2021-just,tam2021improving} have shown promising results when trained in a few-shot setting. They bridge the gap between downstream tasks like text classification and the pre-training of models by converting the dataset into a cloze-style question format that brings it closer to the masked language modelling objective. Using this technique, models with few hundred million parameters can outperform parameter-rich models such as \mbox{GPT-3}~\citep{brown2020language} on various benchmark tasks~\citep{wang-etal-2018-glue} by fine-tuning on just 32 examples.\footnote{This comparison is not entirely fair as the GPT model uses priming and is not fine-tuned on any task-specific data.} Our motivation for adopting this framework is threefold: \emph{(i)}~there has not been much prior work that puts these models under scrutiny in a cross-lingual setting, \emph{(ii)}~often, there is a data scarcity for many languages, which is also the case with stance datasets (only three of our datasets contain more than 2,000 training examples, see Section~\ref{sec:datasets}), \emph{(iii)}~the label inventories of different datasets are often shared or contain synonymous words such as \emph{`pro', 'in favour', 'support'}, etc., which can be strong indicators for the model in both few-shot or full-resource setting.%

\subsection{Cross-lingual Stance Pattern Training}
\label{subsec:method_crossling_stance}

Figure~\ref{fig:approach} shows the architecture of our model. First, we use a simplified PET with a pre-trained language model to predict the likelihood of each label 
to fill a special mask token in a sentence-based template (see \emph{Prompt} below). To obtain a suitable representation (\emph{label embeddings}) for the labels, we use a \emph{`label encoder`} that averages the pooled vectors from the model's token embeddings for each sub-word.
Finally, we take the dot product of the \emph{label embeddings} and the contextualised word embedding for the masked position to obtain the likelihood for each label to fit in.

\paragraph{Prompt} The prompt design is an important aspect of the pattern-exploiting training procedure. In our work, we select a prompt that describes the stance task, rather than a punctuation-based one used in previous work~\citep{schick-schutze-2021-exploiting}. In particular, our prompt is shown below, where the special token changes based on the model choice:

\begin{quote}
    [CLS] \emph{The stance of the following} \textbf{\_\_\_\_\textsuperscript{CONTEXT}} \emph{is} \textbf{[MASK]} \emph{the} \textbf{\_\_\_\_\textsuperscript{TARGET}}.[SEP]
    \label{quote:template}
\end{quote}

Prior work~\citep{qin-eisner-2021-learning,logan2021cutting,lester2021power} has studied %
aspects of PET
such as prompt design, tuning, and selection. Here, we focus on 
the training procedure, and we leave the exploration of these in a multilingual setting for future work.

\paragraph{Label Encoder} A well-known challenge in PET is the need for a fixed number of positions for the label, e.g.,~a single mask is needed for words present in the dictionary such as \emph{`Yes/No'}; however, we need multiple positions to 
predict more complex ones with multiple tokens such as \emph{`Unrelated'}. 
Moreover, if different labels have different lengths, the model needs to ignore some of the positions, e.g.,~to predict a padding inside the sentence. 
The label inventory commonly contains words tokenised into multiple tokens. \citet{schick-schutze-2021-exploiting} propose a simple verbalisation technique where the original labels are replaced with words that can be represented with a single token from the vocabulary, e.g.,~\emph{`Favour'} $\xrightarrow{}$ \emph{`Yes'}, \emph{`Against'} $\xrightarrow{}$ \emph{`No'}. Another possibility is to automatically detect such words, but this yields notable drop in performance compared to manual verbalisation by a domain expert~\citep{schick-etal-2020-automatically}.

Here, we propose a simple, yet effective, approach to overcome this problem. Instead of using a single token representation per label, we take the original label inventory and we tokenise all words, as shown in Figure~\ref{fig:approach}. In the \emph{`Label inventory'} box, we see four labels common for stance tasks and their tokens (obtained by the XLM-R's tokenizer) -- \emph{\{`\_against'\}}, \emph{\{`\_discuss', `ing'\}}, \emph{\{`\_in', `\_favour'\}}, and \emph{\{`\_un', `related', `\_to'\}}. For each token of a label, we extract the vector representation from the MLM pre-trained model's (e.g.,~\XLMR) token embeddings $v^{L_t}_{TE} = TokEmb(L_t)$. Afterwards, we obtain the final label representation ($LE_L$) using an element-wise averaging for all $v^{L_t}_{TE}$ (see Eq.~\ref{eq:tokemb}).

\begin{equation}
    LE_L = \frac{1}{N} \sum_{t=0}^N{TokEmb(L_t)}; \forall_{L} \in \{Labels\}
    \label{eq:tokemb}
\end{equation}

Note that for single tokens, this method defaults to the original MLM task used in learning BERT-based models~\citep{devlin2019bert,liu2019roberta}. The technique of averaging the embedding is shown to be effective with non-contextualised language models such as word2vec~\cite{mikolov2013efficient} and GloVe~\cite{pennington-etal-2014-glove} for representing entire documents or for obtaining a token-level representation with fastText~\citep{joulin-etal-2017-bag}.

Finally, to obtain the label for each example, we take the dot product between the MLM representation for the masked token position, and each of the $LE_L$ vectors. There is no need for padding, as both representations are of the same dimensionality by design ~\citep{conneau2020-xlm-roberta}. Here, we must note that we select the candidates only from the task-related labels; however, we treat the task as a multi-label one, as we describe in more detail below.

\paragraph{Training Objective} We use a standard binary-cross entropy (BCE) loss for each label, where for positive examples, we propagate 1, and for negative ones, we propagate 0. We do not use the original MLM cross-entropy over the entire dictionary, as this will force the model to recognise only certain words as the correct labels, whereas their synonyms are also a valid choice. Moreover, such a loss will prevent further knowledge transfer between tasks and will degraded the model's ability to perform in a zero-shot setting.

\begin{gather}
    \begin{medsize}
    \displaystyle \mathcal{L}_{LE} = 
    \sum_{y^{\prime}_\in y^{p}}{\text{BCE}(p(y^{\prime}|x), 1)} + 
    \sum_{y^{\prime\prime} \in y^{n}}{\text{BCE}(p(y^{\prime\prime}|x), 0)} 
    \end{medsize} \\
    \mathcal{L} = \lambda \cdot \mathcal{L}_{LE} + (1-\lambda) \cdot \mathcal{L}_{MLM}
\end{gather}

\begin{table*}[t]
    \centering
    \setlength{\tabcolsep}{5pt}
    \resizebox{1.00\textwidth}{!}{%
    \begin{tabular}{rl|lllrrl}
	\toprule
	{} &         \bf{Dataset} & \bf{Language} &                                       \bf{Target} &  \bf{Context} & \bf{\#Targets} & \bf{\#Contexts} &                                                           \bf{Labels} \\
	\midrule
	1  &             ans &   Arabic &                                        Headline &    Headline &    2,749 &     2,857 &                          agree (34\%), disagree (63\%), other (2\%) \\
	2  &        arabicfc &   Arabic &                                        Claim &  Article &      421 &     2,897 &       \makecell[lt]{unrelated (68\%), agree (16\%), \\ discuss (13\%), disagree (3\%) } \\
	3  &      conref-ita &  Italian &                                        Tweet &    Tweet &      947 &       963 &                           against (70\%), favor (17\%), none (12\%) \\
	4  &           czech &    Czech &    Smoke ban,  Milos Zeman &  Comment &        2 &     1,455 &                        against (29\%), in favor (24\%), none (48\%) \\
	5  &            dast &   Danish &                               Claim or Topic &     Post &       33 &     2,997 &  \makecell[lt]{commenting (78\%), denying (10\%), \\ querying (3\%), supporting (9\%)} \\
	6  &           e-fra &   French &               \makecell[lt]{Emmanuel Macron, \\ Marine Le Pen} &    Tweet &        2 &     1,112 &                          against (69\%), favour (14\%), none (17\%) \\
	7  &           hindi$^*$ &  Hindi-En &                                    Notebandi &    Tweet &        1 &     3,545 &                           none (55\%), favor (27\%), against (18\%)  \\
	8  &  ibereval-ca &  Catalan &                   Independència de Catalunya &    Tweet &        1 &     4,319 &                          favor (61\%), neutral (36\%), against (3\%) \\
	9  &  ibereval-es &  Spanish &                    Independencia de Cataluña &    Tweet &        1 &     4,319 &                         neutral (59\%), against (33\%), favor (8\%) \\
	10 &           nlpcc$^\ddagger$ &  Chinese &  \makecell[lt]{Two Children, Firecrackers,  \\ IphoneSE, Russia in Syria, \\ Motorcycles ban }&     Post &        5 &     2,966 &                           against (40\%), favor (39\%), none (20\%) \\
	11 &           r-ita &  Italian &                    Referendum costituzionale &    Tweet &        1 &       833 &                           against (58\%), none (22\%), favor (20\%) \\
	12 &        rustance &  Russian &                               Claim or Tweet &  Comment &       17 &       956 &              \makecell[lt]{comment (69\%), query (20\%), \\ support (6\%),  deny (5\%) } \\
	13 &     sardistance &  Italian &                      Movimento delle sardine &    Tweet &        1 &     3,242 &                           against (55\%), favor (24\%), none (21\%) \\
	14 &      xstance-de &   German &                                     Question &   Answer &      173 &    46,723 &                                       against (50\%), favor (50\%) \\
	15 &      xstance-fr &   French &                                     Question &   Answer &      178 &    16,309 &                                       favor (53\%), against (47\%)  \\
	\bottomrule
	\end{tabular}
    }
    \caption{The cross-lingual datasets included in our work and their characteristics.
    If a dataset contains a small number of targets, then we list them, as they are in the dataset.  $^\ddagger$The targets of the \emph{nlpcc} are in Chinese, except \emph{IphoneSE}, we show the respective English translations. $^*$The texts in the \emph{hindi} dataset are code-mixed (Hindi-English).}
    \label{tab:datasets}
\end{table*}

\paragraph{Positive and Negative Sampling}
The label encoder allows for sampling of positive and negative examples at training. This can be useful for tasks such as stance detection, where label inventories can differ, but labels overlap semantically. Indeed, this holds for our datasets, as is apparent in Table ~\ref{tab:datasets} where we see semantically similar labels like \emph{support}, \emph{agree}, \emph{favor} etc. across several datasets.

To obtain a set of synonyms for each label, we use two publicly available sources: \emph{(i)}~Google Dictionary suggestions\footnote{\url{https://github.com/meetDeveloper/freeDictionaryAPI}}; and \emph{(ii)} synsets of the English WordNet~\citep{miller1998wordnet}. However, this is prone to noise, as a word can have multiple meanings, and building a high-quality lexicon would require a human annotator proficient in the target language. 
Thus, we use negative sampling, as unrelated words are also undesirable to predict by the model, rather than using these examples to enrich the positive labels lexicon.

\subsection{Sentiment-Based Stance Pre-Training}
\label{sec:senti:stance}

We propose a novel semi-supervised method for pre-training stance detection models using annotations from a sentiment analysis model. This is motivated by the observation that these are two
closely related tasks (the difference being that sentiment analysis does not have a target).\footnote{Note that we consider the basic, untargeted variant of sentiment analysis here, as more resources exist for it.} %
To illustrate this, consider the sentence \emph{`I am so happy that Donald Trump lost the election.'}, which has a \emph{positive} sentiment, but when expressed towards a specific target, e.g.,~\emph{`Donald Trump'}, then the expected label should be the opposite -- \emph{negative}, or more precisely \emph{against}. This requires for the introduction of targets that can change the sentiment label. For further details how we produce corresponding datasets see Section~\ref{sec:datasets:wikisenti}.

We hypothesise that such pre-training could help bootstrap the model's performance, especially in a low-resource setting, similarly to pre-training on cross-domain stance datasets. We use the same model and pattern as for fine-tuning the cross-lingual stance models, and we use a masked language modelling objective and negative sampling to improve the language model's performance on one hand, and, on the other hand, to allow the model to associate synonyms as the label inventories are very diverse (see Table~\ref{tab:datasets}). We do not do positive sampling as it requires high-quality synonyms, which can only be obtained by manual annotations, while our goal is to design an end-to-end pipeline without a need for human interaction.

\section{Datasets}
\label{sec:datasets}

We use three types of datasets: 15 cross-lingual stance datasets (see Table~\ref{tab:datasets}), English stance datasets, and raw Wikipedia data automatically annotated for stance.
The cross-lingual ones are used for fine-tuning and evaluation, whereas the rest are only used for pre-training.
Appendix~\ref{sec:appendix:dataset} provides additional examples for the cross-lingual datasets shown in Table~\ref{tab:examples}. Further quantitative analysis of the texts are also shown in the Appendix in Table~\ref{tab:tokens_words} and Figure~\ref{fig:tsne_datasets}.

\subsection{Cross-Lingual Stance Datasets}

\textbf{ans}~\citep{khouja-2020-stance}. The Arabic News Stance corpus has paraphrased or contradicting news titles from several major news sources in the Middle East. \\
\textbf{arabicfc}~\citep{baly-etal-2018-integrating} consists of claim-document pairs with true and false claims extracted from a news outlet and a fact checking website respectively. Topics include the \textit{Syrian War} and other related Middle Eastern issues.\\
\textbf{conref-ita}~\citep{lai2018conref} contains tweet-retweet-reply triplets along with their stance annotation pertaining to a polarising \emph{referendum} held in Italy in December 2016 to amend the constitution.\\
\textbf{czech}~\citep{hercig2017detecting} provides stance-annotated comments on a news server in Czech on a proposed \textit{Smoking ban in restaurants} and the Czech president \emph{Miloš Zeman}.\\
\textbf{dast}~\citep{lillie-etal-2019-joint} %
includes stance annotations towards submissions on Danish subreddits covering various political topics. \\
\textbf{e-fra, r-ita}~\citep{LAI2020101075:FRAITA}
consists of French tweets about the 2017 French presidential election and Italian ones about the 2016 Italian constitutional referendum.\\
\textbf{hindi}~\citep{Swami2018AnEC} has Hindi-English code-mixed tweets and their stance towards \emph{demonetisation} of the Indian currency that took place in 2016.\\
\textbf{ibereval}~\citep{taule2017overview} contains tweets in Spanish and Catalan about the \textit{Independence of Catalonia}, collected as part of a shared task held at IberEval 2017.\\
\textbf{nlpcc}~\citep{xu2016nlpcc} contains posts from the Chinese micro-blogging site Sina Weibo about manually selected topics like the \emph{iPhoneSE} or the \emph{open second child policy}.\\  
\textbf{rustance}~\citep{lozhnikov2020rustance} includes posts on Twitter and Russian-focused media outlets on topics relating to Russian politics. The extraction was done in 2017.\\
\textbf{sardistance}~\citep{cignarella2020sardistance} includes textual and contextual information about tweets relating to the Sardines movement%
in Italy towards the end of 2019.\\
\textbf{xstance}~\citep{vamvas2020xstance} contains questions about topics relating to Swiss politics, answered by Swiss political candidates in French, Swiss German, or Italian, during elections held between 2011 and 2020.\\

\subsection{English Stance Datasets}

We use 16 English stance datasets from two recent large-scale studies of multi-task/multi-dataset stance detection~\citep{schiller2021stance,hardalov-etal-2021-cross}. We followed the data preparation and the data pre-processing described in the aforementioned papers as is. The combined dataset contains more than 250K examples, 154K of which are used for training. The data comes from social media, news websites, debating forums, political debates, encyclopedias, and Web search engines, etc. The label inventory includes 24 unique labels. We refer the interested reader to the respective papers for further detail.

\subsection{Sentiment-Based Stance Datasets}
\label{sec:datasets:wikisenti}

We use Wikipedia as a source of candidate examples for constructing our sentiment-based stance dataset due to its size and diversity of topics covered. %
To study the impact language has on pre-training, we construct two datasets: English (\emph{enWiki}) and multilingual (\emph{mWiki}). The latter includes examples from each of the languages covered by some of our datasets. In particular, we use the Wikipedia Python API
to sample random Wiki articles. For the multilingual setup, for each language, we sampled 1,000 unique articles\footnote{We did not include articles in Hindi, as the \emph{hindi} corpus contains texts in Latin, whereas the Wiki articles are in Devanagari.} (non-overlapping among languages), a total of 11,000. For the English-only setup, we sampled the same number of articles.

Next, to obtain the contexts for the datasets, we split the articles (with headings removed) at the sentence-level using a language-specific sentence splitting model from Stanza~\citep{qi-etal-2020-stanza}. Each context is then annotated with sentiment using XLM-T, an \XLMR-based sentiment model trained on Twitter data~\citep{barbieri2021xlm}. We use that model as it covers all the datasets' languages, albeit from a different domain. It produces three labels -- \{\emph{positive, negative, neutral}\}, which we rename to \{\emph{favor, against, discuss}\}, to match the label inventory common for stance tasks. To obtain a target--context pair, we assign a target for each context -- either the `Title' of the article, or, if there is a subheading, the concatenation of the two.
To cover as much as possible of the stance label variety, we also include \emph{unrelated} in the inventory, which we define as \emph{`a piece of text unrelated to the target'}: for this, we randomly match targets and contexts from the existing tuples. The latter class also serves as a regulariser,
preventing 
overfitting to the sentiment analysis task, as it includes examples with positive or negative contexts that are not classified as such. 
The resulting distribution is unrelated (60\%), discuss (23\%), against (10\%), favor (7\%). This aims to match the class imbalance common for stance  tasks~\citep{pomerleau-2017-FNC,baly-etal-2018-integrating,lozhnikov2020rustance}.

Finally, we augment 50\% of the examples by replacing the target (title) with the first sentence from the abstract of the Wikipedia page. These new examples are added to the original dataset, keeping both the original and the augmented ones. Our aim is to also produce long examples such as user posts, descriptions of an events, etc., which are common targets for stance. The resulting dataset contains around 300K examples, which we split into 80\% for training and 10\% for development and testing each, ensuring that sentences from one article are only included in one of the data splits.

\section{Experiments}
\label{sec:experiments}

\paragraph{Models} We evaluate three groups of models: (\emph{i})~without any pre-training, i.e.,~baselines (see next); (\emph{ii})~pre-trained on multiple English stance datasets (\emph{`enstance`}), using automatically labelled instances produced using a sentiment model (\emph{`*Wiki`}), see Section~\ref{sec:senti:stance}; and (\emph{iii})~multi-dataset learning (MDL), i.e.,~we include $N$ examples from each dataset into the training data. We train and evaluate on a single dataset, except in the case of MDL, where we train and evaluate on everything. We choose the best model based on the macro-averaged F1 on all datasets. All models use \XLMRb as their base.

\paragraph{Baselines} In addition to our proposed models (Section~\ref{sec:approach}), we compare to a number of simple baselines:

\textbf{Majority class baseline} calculated from the distributions of the labels in each test set.

\textbf{Random baseline} Each test instance is assigned a target label at random with equal probability.

\textbf{Logistic Regression} A logistic regression trained using TF.IDF word unigrams. The input is the concatenation of separately produced vectors for the target and the context.

\textbf{\XLMR} A conventionally fine-tuned \XLMRb model predicting and back-propagating the errors though the special \textit{$<$s$>$} token.\\

\subsection{Quantitative Analysis}
\label{sec:quantitativeanalysis}

We first analyse the high-level few-shot performance of the proposed models %
using the averaged per-dataset F1 macro. Then, we zoom in on the dataset level and analyse the models in the two most extreme training scenarios: few-shot with 32 examples, and full-resource training.

\paragraph{Few-shot analysis}
\label{sec:few_shotanalysis}

Table~\ref{tab:few_shot_results} shows results for different types of pre-training on top of the pattern-based model (Section~\ref{sec:approach}). The top of the table lists baselines, followed by ablations of training techniques. More fine-grained, performance per dataset is shown in Table~\ref{fig:per_dataset_eval} in Appendix~\ref{sec:appendix:experiments}. We can see that the \emph{`Pattern`} model outperforms random baselines in all shots, except zero. Moreover, there is a steady increase in performance when adding more examples.
The performance saturates 
at around \emph{256} examples, with the difference between it and \emph{all} being 1.3 points F1, whereas in subsequent pairs from previous columns the margin is 3.5 to 5 points. 

\begin{table}[t]
    \centering
    \resizebox{1.00\columnwidth}{!}{%
    \begin{tabular}{l|cccccc}
    \toprule
    & \multicolumn{6}{c}{\bf{Shots}} \\
    \bf{Model} &      \bf{0} &     \bf{32} &     \bf{64} &    \bf{128} &    \bf{256} &    \bf{all} \\
    \midrule
    Majority         & \multicolumn{6}{c}{25.30} \\
    Random         & \multicolumn{6}{c}{30.26}  \\
     Pattern          & 18.25 & 39.17 & 43.79 & 47.16 & 52.15 & 53.43 \\
    \midrule
    & \multicolumn{6}{c}{Pattern + Pre-training} \\
    enWiki           & 28.99 & 45.09 & 47.96 & 50.19 & 53.85 & 54.82 \\
    mWiki            & 28.56 & {45.88} & 48.59 & 51.42 & \underline{54.38} & 57.40 \\
    enstance         & \bf{35.16} & \bf{50.38} & \bf{52.69} & \bf{54.75} & \bf{57.87} & 61.31 \\
    \midrule
    & \multicolumn{6}{c}{Multi-dataset learning} \\
    MDL Pattern      & - & 40.76 & 43.25 & 48.06 & 50.36 &   \underline{61.81} \\
    MDL mWiki        & - & \underline{47.16} & \underline{49.82} & \underline{51.98} & \underline{54.33} &   \bf{62.25} \\
    \bottomrule
    \end{tabular}
    }
    \caption{Few-shot macro-average F1. The random and the majority class baselines use no training, and are constant. \textit{en/mWiki} is a pre-trained on our sentiment-based stance task using English or Multilingual data. \textit{enstance} is pre-trained on all English stance datasets. Multi-dataset learning (\emph{MDL}): we train on $K$ examples from each dataset. }
    \label{tab:few_shot_results}
\end{table}

\begin{table*}[t]
    \centering
    \setlength{\tabcolsep}{6.5pt}
    \resizebox{1.00\textwidth}{!}{%
    \begin{tabular}{l|rrrrrrrrrrrrrrr|r}
    \toprule
        \bf{Model} & \bf{ans} & \bf{arafc} & \bf{con-ita} & \bf{czech} & \bf{dast} & \bf{e-fra} & \bf{hindi} & \bf{iber-ca} & \bf{iber-es} & \bf{nlpcc} & \bf{r-ita} & \bf{rusta.} & \bf{sardi.} & \bf{xsta-de} & \bf{xsta-fr} & \bf{F1\textsubscript{avg}} \\
    \midrule
    Majority  &  26.0 &      20.4 &        27.5 &   22.1 &  21.9 &   28.9 &   23.6 &            25.4 &            24.6 &   19.3 &   24.5 &      19.8 &         26.7 &        33.6 &        35.2 & 25.3 \\
    Random &  24.9 &      20.3 &        26.7 &   33.4 &  17.5 &   25.0 &   32.4 &            28.9 &            31.0 &   32.3 &   31.4 &      20.9 &         28.8 &        50.1 &        50.2 & 30.3 \\
    Logistic Reg. & 31.0 & 32.7 & 31.0 & 29.2 & 21.9 & 33.8 & 33.7 & 45.8 & 39.3 & 29.4 & 60.9 & 24.5 & 32.2 & 62.8 & 64.9 & 38.2 \\
    \XLMRb & 83.2 &      35.7 &        42.3 &   54.7 &  26.2 &   33.0 &   29.3 &            65.9 &            54.2 &   58.2 &   87.6 &      19.8 &         49.9 &        73.2 &        72.7 & 52.4\\
    \midrule
    \multicolumn{16}{c}{Full-resource training} \\
    Pattern   & 84.1 & 39.6 & 34.1 & 48.1 & 34.8 & 34.3 & 43.0 & 67.0 & 56.5 & 51.4 & 79.5 & 32.1 & 49.7 & 73.1 & 74.1 & 53.4 \\
    enWiki    & 86.9 & 38.5 & 42.8 & 50.8 & 25.5 & 48.9 & 45.5 & 65.3 & 57.0 & 51.4 & 88.3 & 22.4 & 50.7 & 73.7 & 74.7 & 54.8 \\
    mWiki     & 83.0 & 40.5 & 63.0 & 55.1 & 32.1 & 49.8 & 45.4 & 68.6 & 57.5 & 54.7 & 93.5 & 32.8 & \bf{52.5} & 64.8 & 67.7 & 57.4 \\
    enstance  & \bf{89.0} & \bf{46.5} & 59.6 & 53.1 & \bf{41.5} & \bf{54.5} & 46.9 & 66.3 & 58.8 & \bf{58.7} & 93.0 & 50.0 & 52.0 & \bf{74.8} & 74.9 & 61.3 \\
    MDL       & 84.7 & 44.8 & \bf{71.7} & 54.1 & 38.2 & 48.9 & 47.5 & \bf{70.5} & 62.1 & 57.3 & 94.1 & \bf{53.0} & 50.3 & {73.9} & \bf{76.1} & 61.8 \\
    MDL mWiki & 82.9 & 42.7 & \bf{71.8} & \bf{56.8} & 40.8 & 49.5 & \bf{48.9} & \bf{70.5} & \bf{64.0} & 58.3 & \bf{96.5} & 51.5 & 49.9 & {73.9} & 75.9 & \bf{62.3} \\
    \midrule
    \multicolumn{16}{c}{Few-shot (32) training} \\
    Pattern   & 38.1 & 26.5 & 31.6 & 43.4 & 25.5 & 40.1 & 35.4 & 39.6 & 35.8 & 37.2 & 54.2 & 44.1 & 37.1 & 47.4 & 51.6 & 39.2 \\
    enWiki    & 39.6 & 33.8 & 46.8 & 44.1 & 27.7 & 47.8 & \underline{39.8} & 46.7 & 39.4 & 45.2 & 75.9 & 31.8 & 41.6 & 58.2 & 58.1 & 45.1 \\
    mWiki     & 45.4 & 32.5 & 46.9 & {46.1} & 26.5 & 50.5 & 39.2 & 42.3 & 40.0 & \underline{47.3} & 80.9 & 31.9 & 43.4 & 57.5 & 57.7 & 45.9 \\
    enstance  & \underline{68.3} & \underline{39.4} & 48.7 & \underline{47.3} & 27.0 & \underline{54.9} & 38.0 & 44.3 & \underline{40.7} & {46.8} & 82.1 & \underline{49.3} & \underline{45.2} & \underline{59.1} & \underline{64.6} & \underline{50.4} \\
    MDL       & 43.7 & 28.4 & 39.8 & 37.8 & \underline{28.3} & 38.7 & 37.7 & 37.5 & 38.6 & 38.9 & 68.0 & 40.9 & 33.8 & 47.2 & 52.1 & 40.8 \\
    MDL mWiki & 47.3 & 31.8 & \underline{58.5} & 44.1 & 27.5 & 47.5 & \underline{39.8} & \underline{48.0} & 39.1 & 46.3 & \underline{82.8} & 35.1 & 44.2 & 57.3 & 58.0 & 47.2 \\
    \bottomrule
    \end{tabular}
    }
    \caption{Per-dataset results with \textit{pre-training}. In multi-dataset learning (MDL), the model is trained on $N$ examples per dataset.}
    \label{tab:per_dataset_full}
\end{table*}

The middle part of Table~\ref{tab:few_shot_results} ablates the stance pre-training on top of the pattern-based model. We first analyse the models trained using the artificial dataset from Wikipedia articles, automatically labelled with a multilingual sentiment model (Section~\ref{sec:senti:stance}). We study the effects of the language of the pre-training data by including two setups  -- \emph{enWiki} that contains only English data, and \emph{mWiki} with equally distributed data among all languages seen in the datasets. Both variants give a sizeable improvement over the baselines in all few-shot settings, especially in low-resource ones. The increase in F1 when using \emph{32} examples is more than 6 points on average; and these positive effects are retained when training on \emph{all} examples. The \emph{mWiki} model outperforms the \emph{Pattern} baseline by 4 points and the \emph{enWiki} -- 1.4 points respectively. The multilingual pre-trained model constantly scores higher than the English pre-trained one. Moreover, we see a tendency for the gap between the two to increase with the number of examples reaching 2.6 points in \emph{all}. 

For pre-training on English stance data (\emph{enstance}), even with 32 examples, we see a large increase in performance of 11 points absolute over the \emph{Pattern} baseline. This model is also competitive, within 3 points absolute on average, to the baseline trained on the whole dataset. Furthermore, the \emph{enstance} model outperforms the pre-training with automatically labelled stance instances (\emph{en/mWiki}). Nevertheless, the \emph{en/mWiki} models stay within 3-5 points F1 in the all shots. The gap in performance is expected, as the \emph{enstance} model is exposed to multiple stance definitions during its extensive pre-training, in contrast to the single one in the \emph{Wiki} and its noisy labels. Finally, only \emph{enstance} surpasses the random baselines even in the zero-shot setting, scoring 35.16 F1, demonstrating the difficulty of this task. We offer additional analyses of the zero-shot performance in Appendix~\ref{sec:zeroshot_analysis}.

The bottom part of Table~\ref{tab:few_shot_results} shows results for multi-dataset learning (\emph{MDL}). Here, the models are trained on $N$ examples from each dataset, instead of $N$ from a single dataset. 
The first row presents the \emph{MDL Pattern} model (without any pre-training). Here, we can see that in few-shot setting training on multiple datasets does not bring a significant performance gains compared to using examples from a single dataset. Nonetheless, when all the data is used for training, F1 notably increases, outperforming the English stance model. Furthermore, combining MDL with the multilingual sentiment-based stance pre-training (\emph{MDL mWiki}) yields an even larger increase -- almost 9 points F1 higher than \emph{Pattern}, 5 points better than \emph{mWiki}, and 1 point -- \emph{enstance}. We attribute the weaker performance on few-shot and the strong performance on full-resource learning of the \emph{MDL}-based models to the diversity of the stance definitions and domains of the datasets, i.e.,~\emph{MDL} fails to generalise and overfits the training data samples in the few-shot setting, however when more data is included, it serves as a regularizer, thus the model's score improves. The phenomena is also seen in other studies on English stance~\citep{schiller2021stance,hardalov-etal-2021-cross}. To some extent, the same regularisation effect comes from the pre-training on the artificial stance task, then the model needs to adjust its weights to the new definition, without having to learn the generic stance task from scratch.

\paragraph{Per-dataset analysis}
\label{sec:perdatset_analysis}

Next, we analyse our experiments on the dataset level. In Table~\ref{tab:per_dataset_full} we present a fine-grained evaluation for each dataset covering the two most extreme data regimes that we run our models in: (\emph{i})~full-resource training and (\emph{ii})~few-shot training with 32 examples. We want to emphasise that we do not include state-of-the-art (SOTA) results in Table~\ref{tab:per_dataset_full} as the setup in most previous work differs from ours, e.g.,~the data splits do not match (see Appendix~\ref{sec:data_splits}), or the use different metrics, etc. For more details about the SOTA refer to the Appendix~\ref{sec:appendix:sota}. For completeness, we include two standard strong baseline models, i.e.,~Logistic Regression and a conventionally fine-tuned \XLMR. %
Both baselines are trained on every dataset separately using all of the data available in its training set.

From our results it is clear that even with all data available from training, a model that does not do any pre-training or knowledge transfer such as the \emph{Logistic Regression} struggles with the cross-lingual stance detection tasks. Even though the model surpasses the random baselines, it falls over 14 points F1 short compared to both the \XLMRb and the \emph{Pattern} model. In turn, the \emph{Pattern} model is 1 point better than the \XLMRb outperforming the random baselines on all datasets. Interestingly, the \XLMRb model fails to beat the random baselines on \emph{hindi} and \emph{rustance}. We attribute this to the code-mixed nature of the former, and the small number of training examples (359) in the latter.

To further understand the results of the models bootstrapped with pre-training or multi-dataset learning, we analyse their per-dataset performance next. From Table~\ref{tab:per_dataset_full} we can see that the \emph{MDL} variants achieve the highest results on 8 out of the \DatasetsCount datasets, \emph{enstance} rank best on 6 and a single win is for \emph{mWiki} on \emph{sardistance}. 

Examining the results achieved by the sentiment-based stance pre-training (\emph{en/mWiki}) we see between 7 and 29 points absolute increase in terms of F1 over the \emph{Pattern} baseline for several datasets -- \emph{czech, conref-ita, e-fra} and \emph{r-ita}. A contradiction example are the two datasets \emph{dast} and \emph{rustance}, where we have a notable drop in F1 compared both to the \emph{Pattern} and the \emph{enstance} models. On one hand this can be attributed on the skewed label distribution, especially in the \emph{support(ing)}, \emph{deny} and \emph{querying} classes, on the other hand that also suggests the stance definition in these two datasets is different than the one adopted by us in the \emph{en/mWiki} pre-training. In turn, \emph{enstance} demonstrates a robust performance on all datasets, as it has been pre-trained on a variety of stance detection tasks.

A common characteristic uniting the datasets, where the \emph{MDL} models achieve the highest F1, is the presence of at least one other dataset with similar topic and language: (\emph{i})~\emph{conref-ita} and \emph{r-ita} are both Italian datasets about a referendum, (\emph{ii})~\emph{ibereval} contains tweets about the ``Independence of Catalonia'' in Catalan and Spanish, and (\emph{iii})~\emph{xstance} contains comments by candidates for elections in Switzerland. This suggests that multi-dataset learning is most beneficial when we have similar datasets.

Finally, we analyse the few-shot training with 32 examples. Here, the highest scoring model on 9 out \DatasetsCount is the \emph{enstance} one. This suggests that other models struggle to learn the stance definition from the cross-lingual datasets by learning from just 32 examples.
This phenomena is particularly noticeable in datasets having a skewed label distribution with one or more of the classes being a small proportion of the dataset such as the two Arabic datasets (\emph{ans} -- other (2\%), \emph{arabicfc} -- disagree (3\%)). Nevertheless, \emph{en/mWiki} models show steady sizeable improvements of 6 points F1 on average on all dataset. On the other hand, as in the full-resource setting, training on multiple datasets (\emph{MDL mWiki}) boosts the performance of \emph{conref-ita} and \emph{r-ita} with 27 points F1 compared to the \emph{Pattern} baseline. However, we must note that this holds true only when we pre-training on a stance task, as the \emph{MDL} model has lower F1. That again is an argument in favour of our hypothesises: (\emph{i})~few-shot training on multiple stance datasets fails to generalise, and (\emph{ii})~combining datasets that cover the same topic and are in the language have the largest impact on the model's score.

\section{Discussion}
\label{sec:discussion}

Our fine-tuning with few instances improves over random and non-neural baselines such as Logistic Regression trained on all-shots, even by more than 20 points F1 on average when training on just 32 instances. However, such models, especially when trained on very few examples, suffer from large variance and instability. In particular, for cross-lingual stance, the pattern-based model's standard deviation ($\sigma$) varies from 1.1 (\emph{conref-ita, nlpcc}) to 8.9 (\emph{ibereval-ca}), with 3.5 on average when trained on 32 examples. Pre-training improves stability by reducing the variance, e.g.,~\emph{en/mWiki} have a $\sigma$ 2.7 with minimum under 1, which is more than 5\% relative change even when comparing to the highest F1 average achieved with 32 examples. The lowest $\sigma$ is when the model is trained on all-shots, and especially in the \emph{MDL} models with 1.7, and 1.4 for the \emph{mWiki} variant.

This variability can be attributed to the known instabilities of large pre-trained language models~\citep{mosbach2021on}, but this does not explain it all. Choosing a right set of data points is another extremely important factor in few-shot learning that calls for better selection of training data for pre-training and fine-tuning~\citep{axelrod-etal-2011-domain,ruder-plank-2017-learning}.

Another important factor is the inconsistency of the tasks in the training data. This is 
visible from our \emph{MDL} experiments, i.e.,~the tasks 
use a variety of 
definitions and labels. Even with more examples in the training set in comparison to single-task training (\DatasetsCount{x}$N$ examples), the models tend to overfit and struggle to generalise to the testing data. In turn, when sufficient resources are available, \emph{MDL} yields sizeable improvements even without additional pre-training.

Having access to noisy sentiment-based stance data in the same languages helps, but transferring knowledge from a resource-rich language (e.g.,~English) on the same task (or set of task definitions) is even more beneficial, in contrast to the data's (see Section~\ref{sec:quantitativeanalysis}) 
and label's language (see  Appendix~\ref{sec:appendix:ablations}). Moreover, when using noisy labels from an external model, there is always a risk of introducing additional bias due the training data and discrepancies in the task definition~\citep{waseem2021disembodied,bender2021parrots}. We observed this for both the \emph{dast} and the \emph{rustance} datasets.

\section{Related Work}

\paragraph{Stance Detection}
\label{sec:relatedwork}

Recent studies on stance detection have shown that mixing {cross-domain} English data improves accuracy and robustness~\citep{schiller2021stance,hardalov-etal-2021-cross}. %
They also indicated important challenges of cross-domain setups such as differences in stance definitions, annotation guidelines, and label inventories. %
Our \emph{cross-lingual} setup adds two more challenges: \emph{(i)}~data scarcity in the target language, which requires learning from few examples, and \emph{(ii)}~need for better multilingual models with the ability for cross-lingual knowledge transfer.

\paragraph{Cross-Lingual Stance Detection}
There have been many efforts to develop multilingual stance systems~\citep{zotova-etal-2020-multilingual,taule2017overview, vamvas2020xstance,agerri2021vaxxstance}.
However, most of them consider 2--3 languages, often from the same language family,
thus only providing limited evidence for the potential of cross-lingual stance models to generalise across languages.
A notable exception is~\citet{LAI2020101075:FRAITA}, who work with 5 languages, but restrict their study to a single family of non-English languages and their domain to political topics only. Our work, on the other hand, spans 6 language families and multiple domains from news~\citep{khouja-2020-stance} to finance~\citep{vamvas2020xstance}.

\paragraph{Stance and Sentiment}
Sentiment Analysis has a long history of association with stance~\citep{somasundaran-etal-2007-detecting, somasundaran-wiebe-2010-poldeb}. %
Sentiment is often annotated in parallel to stance~\citep{mohammad-etal-2016-semeval,hercig2018_stance_sentiment} and has been used extensively as a feature~\citep{ebrahimi-etal-2016-joint,sobhani-etal-2016-detecting,sun-etal-2018-stance} or as an auxiliary task~\citep{li-caragea-2019-multi, sun2019stance} for improving stance detection. Missing from these studies, however, is leveraging sentiment annotations to generate noisy stance examples,
which we explore here: for English and in a multilingual setting.

\paragraph{Pattern-based Training}

Recently, prompt or pattern-based training has emerged as an effective way of exploiting pre-trained language models for different tasks in few-shot settings~\citep{petroni-etal-2019-language,schick-schutze-2021-exploiting}. \citet{brown2020language} introduced a large language model (i.e.,~\mbox{GPT-3}), which showed strong performance on several tasks through {demonstrations} of the task. \citet{schick-schutze-2021-exploiting, schick-schutze-2021-just} proposed Pattern-Exploiting Training (PET), a novel approach using comparatively smaller masked language models through Cloze-style probing with task informed patterns. \citet{tam2021improving} build on PET, with an additional loss that allows them to circumvent the reliance on unsupervised data and ensembling.
There have been studies on aspects of prompt-based methods such as performance in the absence of prompts~\citep{logan2021cutting}, quantifying scale efficiency~\citep{le-scao-rush-2021-many}, learned continuous prompts~\citep{li-liang-2021-prefix,lester2021power,qin-eisner-2021-learning} or gradient-based generated discrete prompts~\citep{shin-etal-2020-autoprompt}. 
\citet{liu2021prompt_survey} offer a survey of prompt-based techniques. %
We focus on the few-shot advantages offered by PET methods and evaluating them in a cross-lingual setting.

\section{Conclusion and Future Work}
\label{sec:conclusions}

We have presented a holistic study of cross-lingual stance detection. We investigated PET with different (pre-)training procedures and extended it by proposing a novel label encoder that mitigates the need for translating labels into a single verbalisation. In addition to that, we introduced a novel methodology to produce artificial stance examples using a set of sentiment annotations. This yields sizeable improvements on \DatasetsCount datasets over strong baselines, i.e.,~more than 6\% F1 absolute in a low-shot and 4\% F1 in a full-resource scenario. Finally, we study the impact of multi-dataset learning and pre-training with English stance data, which further boost the performance by 5\% F1 absolute.

In future work, we plan to experiment with more sentiment-based models and stance task formulations, as well as different prompt engineering techniques.

\bibliography{bibliography}

\appendix

\section{Fine-Tuning and Hyper-Parameters}
\label{sec:appendix:hyperparams}

\begin{itemize}
    \item All the neural-based models (i.e.,~\XLMR and its variants) are developed in Python using PyTorch~\citep{NEURIPS2019_9015} and the Transformers library~\citep{wolf-etal-2020-transformers}.
    \item Logistic regression and TF.IDF are implemented using Scikit-learn~\cite{JMLR:v12:pedregosa11a}.
    \item All neural-based models are trained for 5 epochs with batch size 16, using the AdamW optimiser~\citep{loshchilov2018decoupled}, with weight decay 1e-8, $\beta_1$ 0.9, $\beta_2$ 0.999, $\epsilon$ 1e-08.
    \item All models use class weights,
    in BCE only the positive class is weighted.
    \item We use a maximum sequence length of 150 tokens. Afterwards, we truncate only the target and the context sequences token by token, keeping the tokens from the pattern unchanged, starting from the longest sequence in the pair.
    \item For logistic regression we convert the text to lowercase, and limit the dictionary in the TF.IDF to 15,000 uni-grams. The vocabulary is fit on the concatenated target and context. The two are then transformed separately to form the individual vectors.
    \item All the hyper-parameters are tuned on the corresponding validation sets. 
    \item In models, trained on a single dataset, we choose the best checkpoint based on its performance on the corresponding validation set. For multi-dataset training, we choose the checkpoint that performs best on the combined validation sets. The models are evaluated on every 200 steps.
    \item We run each setup three times and then average the scores. In few-shot setting, we use three different subsets from the training dataset with $K$ instances, while keeping the same seed for the models, whereas in full-resource setting, we set different seeds for each run.
    \item Each single dataset experiment took roughly 20 minutes on a single NVIDIA V100 GPU using half precision.
\end{itemize}

\paragraph{Zero-shot} For our zero-shot experiments, we only pass the labels from the corresponding task's inventory to the models to obtain a ranking, we then take the label with maximum probability. 

\paragraph{Few-shot} The few-shot models are trained w/ LR 1e-05 and warmup 0.06. The number of training steps is: (\emph{i})~2,000 steps for single dataset training, (\emph{ii})~4,000 steps for multi-dataset learning.

\paragraph{Full-resource} The \XLMR, both \emph{$<$s$>$} fine-tuned and pattern-based, models are trained w/ LR 3e-05 and warmup 0.06 for 8 epochs. The logistic regression model is trained using the Limited-memory BFGS until convergence.

\paragraph{Sentiment-Based Pre-Training} We train for 3 epochs with MLM of 12.5\% and 2 negative samples per-label.

\section{Dataset}
\label{sec:appendix:dataset}

\begin{figure}[t]
    \centering
    \includegraphics[width=\columnwidth]{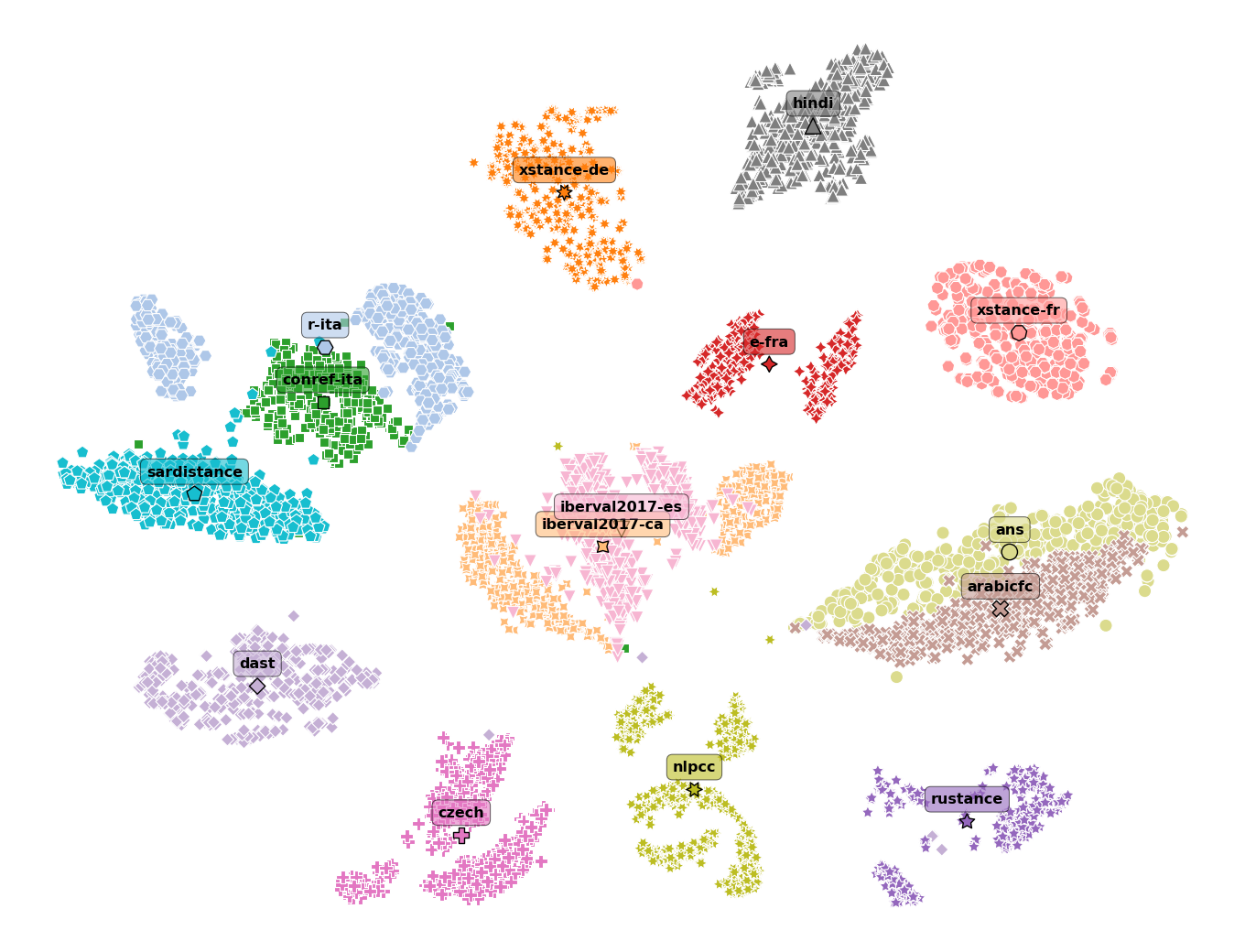}
    \caption{tSNE plot of the datasets using the representations from the \emph{$<$s$>$} token (\XLMRb). The highlighted points denote dataset centroids.}
    \label{fig:tsne_datasets}
\end{figure}

\begin{table}[t]
    \centering
    \small
    \begin{tabular}{l|rrrr}
    \toprule
            Dataset &   Train &    Dev &    Test &   Total \\
    \midrule
                ans &   2,652 &    755 &     379 &   3,786 \\
           arabicfc &   1,140 &    381 &   1,521 &   3,042 \\
         conref-ita &     361 &    121 &     481 &     963 \\
              czech &     547 &    183 &     730 &   1,460 \\
               dast &   1,127 &    376 &   1,504 &   3,007 \\
              e-fra &     318 &    106 &     692 &   1,116 \\
              hindi &   1,329 &    444 &   1,772 &   3,545 \\
        ibereval-ca &   1,619 &    540 &   2,160 &   4,319 \\
        ibereval-es &   1,619 &    540 &   2,160 &   4,319 \\
              nlpcc &   1,119 &    374 &   1,493 &   2,986 \\
              r-ita &     500 &    166 &     167 &     833 \\
           rustance &     359 &    120 &     479 &     958 \\
        sardistance &   1,599 &    533 &   1,110 &   3,242 \\
         xstance-de &  33,850 &  2,871 &  11,891 &  48,612 \\
         xstance-fr &  11,790 &  1,055 &   4,368 &  17,213 \\
         \midrule
              Total &  59,929 &  8,565 &  30,907 &  99,401 \\
    \bottomrule
    \end{tabular}
    \caption{Number of examples for each dataset.}
    \label{tab:data_splits}
\end{table}

In this section, we first discuss the datasets characteristics in terms of data splits, then we present the word-level statistics and finally analyse the transformed labels after tokenisation in terms of sentence pieces~\citep{kudo-richardson-2018-sentencepiece}.

In order understand and demonstrate the relationship between the tasks, we proportionally sample 25,000 instances from each dataset and pass them though the \emph{$<$s$>$} token of a \XLMRb model to obtain a sentence-level representation of each. The input has the following form: \texttt{$<$s$>$} context \texttt{$<$/s$>$} target. The resulting clusters are shown in Figure~\ref{fig:tsne_datasets}. As the model is not fine-tuned on any down-stream task, we see that each cluster covers only one language, e.g.,~\emph{ans} and \emph{arabicfc}, both in Arabic, are grouped in the right part of the plot, Spanish/Catalan (\emph{ibereval}) are in the middle, and Italian (\emph{conref-ita, r-ita, sardistance}) are on the left. Surprisingly, the two French ones --- \emph{xstance-fr} and \emph{e-fra}) --- are well-separated. We attribute this to their different domains, i.e.,~tweets about French politicians vs. stance of  political candidates towards Swiss policies. Moreover, even datasets from the same source domain such as \emph{xstance-de} and \emph{xstance-fr} do not have any overlapping points. On the other hand, we see that the centroids of datasets in the same language and with related topics such as \emph{ibereval-ca} and \emph{ibereval-es} or \emph{conref-it} and \emph{r-ita} are also close in the latent space. 

\subsection{Data Splits}
\label{sec:data_splits}

Table~\ref{tab:data_splits} shows the number of examples in the training, development, and test datasets. Due to their small size, some of the datasets did not provide any standard splits as their authors used cross-validation. The datasets that did provide such splits are \emph{ans, e-fra, r-ita, xstance}. For the rest of the datasets, we produced our own splits with ratio 3:1:4 for the training, development, and testing, respectively. \emph{sardistance} had only train/test split, and thus we further split the training set into train/dev using a ratio of 3:1. Finally, we used only the training set of \emph{ibereval}, and Task A's training data from \emph{nlpcc}, as the testing data was not publicly available. 

We generated the few-shot subsets by randomly sampling $K$ examples from the training set. We allow example overlap between the 3 subsets, but we enforce the subset to have at least one instance from each class. We think that this setup is more realistic and recreates a real-world scenario, in contrast to sampling an equal number of examples per-class as is often done in previous work on few-shot learning.

\begin{table}[t]
    \centering
    \resizebox{1.00\columnwidth}{!}{%
    \begin{tabular}{l|rrrr}
    \toprule
    \bf{Dataset} &  \bf{Tokens (Words)} & \multicolumn{3}{c}{\bf{Tokens}} \\
    {} &             \makecell[c]{Mean} &  25\% & Median &     Max \\
    \midrule
    ans            &      27.0 (18.4) &   22 &     26 &      58 \\
    arabicfc       &  2185.5 (1638.5) &  627 &  1,299 &  10,370 \\
    conref-ita     &     147.9 (74.2) &  134 &    149 &     236 \\
    czech          &      53.0 (33.4) &   26 &     40 &     326 \\
    dast           &      87.5 (64.9) &   37 &     60 &   1,124 \\
    e-fra          &      41.1 (24.1) &   35 &     42 &      66 \\
    hindi          &      38.8 (21.8) &   31 &     40 &     105 \\
    ibereval-ca    &      40.8 (20.7) &   33 &     42 &      83 \\
    ibereval-es    &      38.7 (22.8) &   32 &     39 &      74 \\
    nlpcc          &      59.7 (52.9) &   33 &     56 &     215 \\
    r-ita          &      48.0 (17.7) &   43 &     49 &      69 \\
    rustance       &      72.7 (32.1) &   62 &     71 &     210 \\
    sardistance    &      51.4 (36.9) &   34 &     50 &     130 \\
    xstance-de     &      68.2 (46.5) &   49 &     63 &     198 \\
    xstance-fr     &      80.1 (61.5) &   56 &     74 &     230 \\
    \bottomrule
    \end{tabular}
    }
    \caption{Statistics about the words for each dataset (using {\XLMR}’s tokeniser). Numbers in parenthesis show word count using Stanza tokeniser for the dataset's language.}
    \label{tab:tokens_words}
\end{table}

\subsection{Word Statistics}

Next, in Table~\ref{tab:tokens_words} we present the statistics in terms of number of words/tokens per dataset. The first column shows the average number of words (tokens). The statistics are obtained using language-specific tokenisers from Stanza~\citep{qi-etal-2020-stanza}. In turn, the tokens statistics are calculated in terms of sentence pieces after applying \XLMR's tokeniser. In subsequent columns, we describe the text lengths' distribution by reporting the 25th and 50th percentile, and the max.

Most of the datasets contain short texts, i.e.,~under 100 words/pieces, expect for \emph{arabicfc} that is built using retrieved documents from the Google search engine, and \emph{conref-ita} whose contexts are triplets of tweets (tweet-retweet-reply).

\begin{table}[t]
    \centering
    \begin{tabular}{ll}
    \toprule
    {Label} &           Tokenised w/ \XLMR \\
    \midrule
    querying         &      ['\_que', 'ry', 'ing'] \\
    argument for     &      ['\_argument', '\_for'] \\
    negative         &              ['\_negative'] \\
    against          &               ['\_against'] \\
    con              &                   ['\_con'] \\
    for              &                   ['\_for'] \\
    commenting       &        ['\_comment', 'ing'] \\
    unrelated        &         ['\_un', 'related'] \\
    other            &                 ['\_other'] \\
    none             &              ['\_non', 'e'] \\
    neutral          &               ['\_neutral'] \\
    deny             &              ['\_de', 'ny'] \\
    agree            &                 ['\_agree'] \\
    observing        &         ['\_observ', 'ing'] \\
    denying          &           ['\_den', 'ying'] \\
    discuss          &               ['\_discuss'] \\
    pro              &                   ['\_pro'] \\
    supporting       &        ['\_support', 'ing'] \\
    endorse          &       ['\_en', 'dor', 'se'] \\
    undermine        &         ['\_under', 'mine'] \\
    disagree         &              ['\_disagree'] \\
    favor            &                 ['\_favor'] \\
    refute           &            ['\_refu', 'te'] \\
    in favor         &          ['\_in', '\_favor'] \\
    support          &               ['\_support'] \\
    favour           &            ['\_f', 'avour'] \\
    anti             &                  ['\_anti'] \\
    argument against &  ['\_argument', '\_against'] \\
    positive         &              ['\_positive'] \\
    comment          &               ['\_comment'] \\
    question         &              ['\_question'] \\
    query            &             ['\_que', 'ry'] \\
    \bottomrule
    \end{tabular}
    \caption{Resulting sub-words for each label after tokenisation with \XLMR's tokeniser.}
    \label{tab:token_labels}
\end{table}

\subsection{Label Tokens}

In Table~\ref{tab:token_labels}, we include the labels and their tokenised forms. Each label is present in at least one stance dataset, either English or multilingual. We include the tokens obtained from \XLMR, as we use it for training the pattern-based models. 

The results confirm that common words in base form such as \emph{positive, negative, neutral} are tokenised into a single token. In turn, rare words such as \emph{endorse}, conjugated verbs (\emph{supporting, querying}), words with prefixes (\emph{unrelated}), British spelling (\emph{favour}), or multi-word expressions (\emph{argument for, in favor}) are broken into multiple pieces.

\begin{table*}[ht!]
    \centering
    \scriptsize
    \setlength{\tabcolsep}{3pt}
    \begin{tabularx}{\textwidth}{lp{3.0cm}Xl}
    \toprule
    \bf{Dataset} & \bf{Target} & \bf{Context} & \bf{Label} \\
    \midrule
    ans & Kuala Lumpur: A Syrian has been stuck at the airport for more than a month & Solve the crisis of the Syrian stuck in Kuala Lumpur airport & disagree \\
    \midrule
    arabicfc & UN Special Envoy for Syria Staffan de Mistura plans to organize a next round of Syrian peace talks in Geneva in the second half of January & The tour came a few days after the end of the Geneva 8 round between the opposition and regime delegations, which suffered a catastrophic failure as a result of the regime's refusal to engage in negotiations, creating pretexts in order to evade political obligations, and it is expected to do the same in order to disrupt the path of Asta... & discuss \\
    \midrule
    conref-ita & Soon live at \#agorarai on Rai Tre. And if these two vote yes ... \#iovotono http://url & [T1] @user \#IOVOTONO \#IOVOTONO \#IOVOTONO \#IOVOTONO \#IOVOTONO  \#IOVOTONO \#IOVOTONO \newline AND YOU MUST GO HOME 
    \newline \newline
    [T2] RT @user: 10 MINUTES to explain why Sunday's vote it will affect our lives. Help me RUN \#iovotono ... & against \\
    \midrule
    czech & MILOŠ ZEMAN & He will make you laugh when you realize where Zeman is dragging us. ;-D & against \\
    \midrule
    dast & Gymnasium students boycott term exam in protest against supervision & No you can choose to write the assignment by hand. & commenting \\
    \midrule
    e-fra & Emmanuel Macron & Michel Onfray: "Macron has seduced all the uneducated, that's a lot of people." - http://url via @Dailymotion & against \\
    \midrule
    hindi & Notebandi & Sir Modi you did a fantastic job for the country with demonetisation. Now please do a surgical on reservation too. It's taking the country backwards. &     favor \\
    \midrule
    iber-ca & Independence of Catalonia & \#27S Brutal! at 5 pm in St.Miquel de Balenyà 71\% of the population have already voted! Never seen! & neutral \\
    \midrule
    iber-es & Independence of Catalonia & The truth is that the only one who can be truly happy tonight is a woman: @user @user \#27S & against \\
    \midrule
    nlpcc & Set off firecrackers during the Spring Festival & [There are fewer firecrackers set off during the Spring Festival, and the air quality in urban areas has improved.] A person in charge of the Municipal Environmental Protection Bureau believes that more and more citizens are aware of the environmental impact of setting off fireworks and firecrackers and have reached a consensus on ``not setting off firecrackers''. Compared with previous years ... & against \\
    \midrule
    r-ita & Constitutional Referendum & \#DiMartedi \#Travaglio: the markets plot do not tremble. To the \#referendumcostituzionale \#IoVotoNO & against \\
    \midrule
    rustance & \#SYRIA  The Russian Ministry of Defense publishes an undeniable confirmation of the United States' provision of combat-ready cover ... http://url & @mod\_russia And no one thought that the United States had access to @mod\_russia accounts. They can publish the devil ... http://url & support \\                                                                                                                
    \midrule
    sardistance & Movement of Sardines & \#King Zingaretti, the people do not have the funds for education! \newline * Oh well, then give them some sardines !! \newline \#Manovra2020 \#instruction \#Fioramonti & against \\
    \midrule
    xstance-de & Do you give priority to tax cuts at the federal level in the next four years? & Yes, because the federal treasury has been making a surplus for years. With small steps so that the effect can be tested. &
   favor \\
    \midrule
    xstance-fr & Should paid paternity leave of several weeks be introduced in addition to the existing maternity insurance? & Switzerland is quite late on this issue and this would help establish a little more equality between men and women! & favor \\
    \bottomrule
    \end{tabularx}
    \caption{Examples from each dataset. The examples are automatically translated to English using Google Translate. Long contexts are trimmed due to length constraints of the table.}
    \label{tab:examples}
\end{table*}

\section{Experiments}
\label{sec:appendix:experiments}

In this section, we cover additional experiments and state-of-the-art results. For the latter, we must emphasise that our setup, and respectively results, are not comparable with most of the previous work due to their data splits not being publicly available or their choice of evaluation metrics (see Appendices~\ref{sec:data_splits} and~\ref{sec:appendix:sota} for more details).

The rest of the section covers the model's performance in different setups: sentiment-based stance task (Appendix~\ref{sec:appendix:sentistance_experiments}), zero-shot inference (Appendix~\ref{sec:zeroshot_analysis}), per-dataset few-shot with 64--256 instances (Appendix~\ref{sec:appendix:few-shot}, and finally an ablation study of prominent model components (Appendix~\ref{sec:appendix:ablations}).

\begin{table*}[t]
    \centering
    \resizebox{1.00\textwidth}{!}{%
    \begin{tabular}{l|rrrrrrrrrrrrrrr|r}
    \toprule
        \bf{Model} & \bf{ans} & \bf{arafc} & \bf{con-ita} & \bf{czech} & \bf{dast} & \bf{e-fra} & \bf{hindi} & \bf{iber-ca} & \bf{iber-es} & \bf{nlpcc} & \bf{r-ita} & \bf{rusta.} & \bf{sardi.} & \bf{xsta-de} & \bf{xsta-fr} & \bf{F1\textsubscript{avg}} \\
    \midrule
    Baseline (Max) & 26.0 & 20.4 & 27.5 & 33.4 & \bf{21.9} & 28.9 & \bf{32.4} & \bf{28.9} & \bf{31.0} & 32.3 & 31.4 & \bf{20.9} & 28.8 & 50.1 & 50.2 & 30.9 \\
    \multicolumn{16}{c}{Zero-shot inference} \\
    Pattern & 26.0 & 5.7 & 27.5 & 14.6 & 4.5 & 28.9 & 10.1 & 1.9 & 16.8 &   19.3 &   24.5 & 2.9 & 26.7 & 33.1 & 31.3 & 18.3 \\
    mWiki    & 25.5 & 14.1 & \bf{35.8} & 18.1 & 14.4 & \bf{34.2} & 24.5 & 21.7 & 16.5 & 35.6 & 33.8 &  7.5 & 38.0 & 49.8 & 58.8 & 28.6  \\
    enstance & \bf{46.2} & \bf{35.5} & 20.2 & \bf{35.6} & 20.3 & 33.3 & 21.4 & 22.3 & 18.4 & \bf{45.8} & \bf{46.6} & 15.3 & \bf{44.9} & \bf{60.7} & \bf{61.1} & \bf{35.2} \\
    \bottomrule
    \end{tabular}
    }
    \caption{Zero-shot results of the models. Baseline (Max) is the maximum per-dataset between the Random and the Majority class baselines.}
    \label{tab:zero_shot}
\end{table*}

\subsection{State-of-the-Art}
\label{sec:appendix:sota}

In Table~\ref{tab:SOTA} we present the previously reported state-of-the-art results on the cross-lingual datasets included in our study. The results for 12 out of 15 datasets, however, are not comparable to the ones we report (see Table~\ref{tab:per_dataset_full}) because of mismatch in the data split and metrics between our and the original setup. As highlighted in Section~\ref{sec:data_splits}, our splits differ from the original ones in 10 out of 15 datasets due to unavailability of public development or test sets, making direct comparison infeasible. Furthermore, for \emph{r-ita} and \emph{e-fra}, the reported F1-macro results in previous work, are only calculated across the \emph{favor} and \emph{against} classes, ignoring the \emph{none} class. We calculate our results across all classes as it more accurately represents a real-world setting, where many text pieces do not take a stance towards a specific target.

\begin{table}[t]
    \centering
    \resizebox{1.00\columnwidth}{!}{%
    \begin{tabular}{l|rcr}
    \toprule
    \bf{Dataset} &    \bf{Score} &    \bf{Comparable} & \bf{Source}\\
    \midrule
    ans & 90.? & \faCheck & \citet{alhindi-etal-2021-arastance} \\
    arabicfc & 52.? & \faTimes & \citet{alhindi-etal-2021-arastance} \\
    conref-ita & 76.? & \faTimes & \citet{lai2018conref} \\
    czech & 51.? & \faTimes & \citet{hercig2017detecting} \\
    dast & 42.? & \faTimes & Lillie et al. (2020) \\%\citet{lillie-etal-2019-joint} \\
    e-fra & 58.5 & \faTimes & \citet{LAI2020101075:FRAITA} \\
    hindi & $^*$58.7 & \faTimes & \citet{Swami2018AnEC}\\
    iber-es & 48.8 & \faTimes & \citet{taule2017overview}\\
    iber-ca & 49.0 & \faTimes & \citet{taule2017overview}\\
    nlpcc & 75.? & \faTimes & \citet{su2021-chinesestance}\\
    r-ita & 95.5 & \faTimes & \citet{LAI2020101075:FRAITA}\\
    rustance & 83.2 & \faTimes & Lozhnikov et al. (2018)
    \\
    sardistance & 68.5 & \faTimes & \citet{cignarella2020sardistance} \\
    xstance-de & 77.9 & \faCheck & Schick et al. (2021a) \\ 
    xstance-fr & 79.0 & \faCheck & Schick et al. (2021a) \\
    \bottomrule
    \end{tabular}
    }
    \caption{Previously reported best results (F1\textsubscript{macro}) on the datasets. We report \faCheck{} when our splits and metric match the original ones, and \faTimes{} when there is a mismatch. Not all datasets report their score with the same precision, therefore we rounded the numbers to the first digit, and added a `\textbf{?}' where the precision was lower. $^*$Only accuracy reported.}
    \label{tab:SOTA}
\end{table}

\subsection{Sentiment-Based Stance}
\label{sec:appendix:sentistance_experiments}

In Table~\ref{tab:wiki:eval} we present the results on the sentiment-based stance pre-training task in terms of F1. In particular, we train each model only on either English or multilingual data (see Section~\ref{sec:datasets:wikisenti}), and respectively evaluate on a testing dataset containing examples in the same language(s). 

For the results shown, we can see that the difference in terms of F1\textsubscript{macro} between the two setups is marginal, under 0.5 points. However, when analysing the fine-grained scores for each label, we observe some larger differences. More specifically, on the \emph{favor} and \emph{against} classes the English model is performing better, whereas the multilingual one scores higher on the \emph{unrelated} instances. The performance on the \emph{discuss} class for both models is more or less equal.

Nevertheless, our experiments show that the multilingual model performs better on the cross-lingual stance datasets (see Section~\ref{sec:quantitativeanalysis}). This suggests that the improvements comes from the language-specific pre-training, not from the model's performance on the auxiliary sentiment-based stance task.

\begin{table}[t]
    \centering
    \setlength{\tabcolsep}{3pt}
    \resizebox{1.00\columnwidth}{!}{%
    \begin{tabular}{l|rrrr|r}
    \toprule
         \bf{Model} & \bf{F1\textsubscript{against}} & \bf{F1\textsubscript{discuss}} & \bf{F1\textsubscript{favor}} & \bf{F1\textsubscript{unrelated}} & \bf{F1\textsubscript{macro}} \\
         \midrule
         Multiling. & 78.20 & \bf{92.29} & 75.77 & \bf{90.82} & 84.27 \\
         English    & \bf{80.03} & \bf{92.42} & \bf{78.29} & 88.19 & \bf{84.73} \\
         \bottomrule
    \end{tabular}
    }
    \caption{The model performance on the sentiment-based stance pre-training task with automatically labelled Wikipedia articles. Each row shows the results on the corresponding English and multilingual testing sets.}
    \label{tab:wiki:eval}
\end{table}

\subsection{Zero-Shot Analysis}
\label{sec:zeroshot_analysis}

In Table~\ref{tab:zero_shot} we present the results for zero-shot setting. We include three variants of the pattern-based models -- one without any pre-training (\emph{Pattern}), one pre-trained on the multilingual stance-based sentiment task (\emph{mWiki}), and finally the model trained on English stance data (\emph{enstance}). 

First, we see that the \emph{Pattern} model has no prior knowledge of how to solve the tasks given just a template. Moreover, the model always predicts a single class, not necessarily the majority one, therefore some of the scores exactly match the F1 from the \emph{Baseline (Max)} row. 

Next, we study the \emph{mWiki} pre-trained model. In general, it does not outperform the baseline in terms of macro-averaged F1, however the model still shows positive results on some of the datasets, i.e.,~\emph{conref-ita, e-fra, nlpcc, r-ita, sardistance} and~\emph{xstance-fr}. These improvements, over the random/majority baselines, are mostly due to the higher F1 scores in two of the labels -- \emph{against} and \emph{favor}. We must note that this increase is not completely related to the naming of the labels as for other datasets, having similar label inventories such as the \emph{hindi} and \emph{ibereval}, all models fail to outperform the baseline. 

Finally, the \emph{enstance} variant successfully surpasses the baseline results on 9 out of the 15 datasets, and in turn achieves 5 points absolute higher F1 than the baseline. On one hand, this indicates the model's potential to perform in a zero-shot setting, however, adding even as little as 32 examples can drastically improve performance (see Tables~\ref{tab:few_shot_results} and~\ref{tab:per_dataset_full}), i.e.,~\emph{enstance} scores 50.4 (32-shots) vs. 35.2 (zero-shot). Moreover, even when we have datasets that share the same label inventories as pre-training tasks, there is no guarantee that the model would be able to transfer the knowledge in zero-shot setting, e.g.,~\emph{support, query, comment, discuss} are present in \emph{dast} and \emph{rustance} from the cross-lingual datasets, and in the English datasets used for  pre-training (e.g.,~the SemEval task: RumourEval~\citep{derczynski-etal-2017-rumoureval}).

\subsection{Per-Dataset Few-Shot Results}
\label{sec:appendix:few-shot}

\begin{figure*}[t]
    \centering
    \includegraphics[height=0.95\textheight]{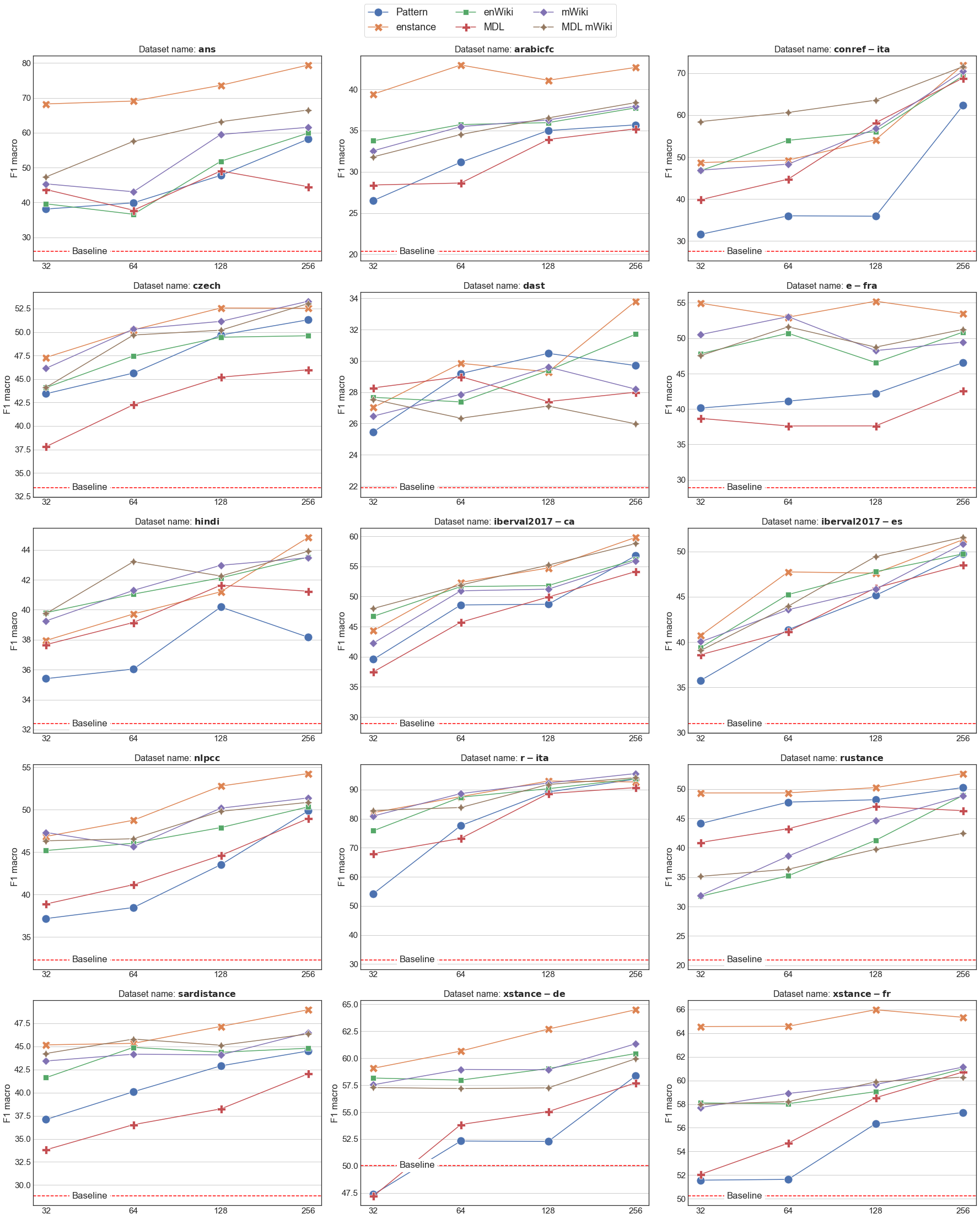}
    \caption{Per-dataset few-shot results. The red dashed line (Baseline) shows the maximum between the random and majority class baseline. The name of the models are shown on top of the figure.}
    \label{fig:per_dataset_eval}
\end{figure*}

Here, we analyse the per-dataset few-shot results shown in Figure~\ref{fig:per_dataset_eval}. Our experiments suggest that even with 32 examples, our models outperform the random/majority baseline on almost all datasets. Only exceptions being the models with no additional pre-training   (\emph{Pattern}, \emph{MDL}), which fail to do so on \emph{xstance-de}. This suggests that the \emph{xstance} task is complex and is not trivial to learn from so few examples. The same models also score close to the baseline on its French version (\emph{xstance-fr}), which in general is easier than the German variant (see Tables~\ref{tab:per_dataset_full} and~\ref{tab:SOTA}), further supports this observation.

Additionally, our experiments show that with increased the training instances, the positive growth in F1 is retained in all datasets. This is also visible from the F1 scores (macro-averaged over all datasets) in Table~\ref{tab:few_shot_results}. Moreover, the \emph{Pattern} model performs considerably worse than the \emph{en/mWiki} pre-trained models and \emph{MDL}, except on \emph{dast}, and \emph{rustance}. This again supports our hypothesis that the latter two datasets adopt different stance definitions compared to the other cross-lingual stance detection tasks. 

Nevertheless, increasing the number of examples does not necessarily lead to an increase in performance in low-shot setting. In particular, for the \emph{enstance} model, we see very small or no gain on several datasets, independent of the number of shots, i.e.,~\emph{arabicfc, rustance}, and \emph{xstance-fr}. Nonetheless, going up to 256 shots often leads to sizeable performance gains in the \emph{Pattern} model by more than 15 points F1 (\emph{ans, conref-ita, ibereval-ca/es, nlpcc}. We attribute this to the large class imbalance that requires the models to be able to learn a target class from one or two instances.

\subsection{Ablations}
\label{sec:appendix:ablations}

In Table~\ref{tab:ablations}, we ablate different components from the pattern training. For a baseline model, we use a \XLMRb fine-tuned using patterns (i.e., \emph{Pattern}). We present the results in 32--256 shots, where on each row we add one additional component to the baseline model. From the first two rows, we can see that adding two negatives, and masking some of the input tokens does not help. We attribute the former to the specificity of the label inventories and the introduced additional imbalance between positive to negative samples. For the latter, \citet{schick-schutze-2021-exploiting} also observed that the MLM loss does not bring additional performance gains when the number of shots increases (but still is beneficial in the low-shot setting). We hypothesise that such a small number of examples are not enough for the model to adapt to the tasks' domains and languages, albeit previous work shows that further MLM pre-training on task-specific data helps~\citep{han-eisenstein-2019-unsupervised}.

Finally, we experiment with translating the label inventories into the corresponding language of the dataset.\footnote{We used Google Translate to obtain the translations of the target labels. We made sure they have the same meaning and form as the originals.} Using these translated labels, we see improvements in terms of F1 -- between 0.36 and 1 point absolute over the \emph{Pattern} baseline. However, with increased number of examples, the improvement diminishes. This suggests that the language of the labels is not a major factor when training the models.

\begin{table}[t!]
    \centering
    \begin{tabular}{l|rrrr}
    \toprule
    \bf{Model} &    \bf{32} &    \bf{64} &   \bf{128} &   \bf{256} \\
    \midrule
    Pattern   & 39.17 & 43.79 & 47.16 & 52.15 \\
    w/ 2 Negatives  & 38.88 & 42.11 & 45.65 & 50.57 \\
    w/ MLM 12.5\%   & 38.37 & 42.92 & 46.79 & 51.60 \\
    w/ Translated Labels & \bf{39.89} & \bf{44.73} & \bf{47.61} & \bf{52.51} \\
    \bottomrule
    \end{tabular}
    \caption{Ablations table.}
    \label{tab:ablations}
\end{table}

\end{document}